\documentclass{article}

\usepackage{multirow}
\usepackage{graphicx}
\usepackage{caption}
\usepackage{subcaption}
\usepackage{bm}
\usepackage[numbers]{natbib} 

\usepackage[final]{neurips_2024}

\usepackage[utf8]{inputenc} 
\usepackage[T1]{fontenc}    
\usepackage{hyperref}       
\usepackage{url}            
\usepackage{booktabs}       
\usepackage{amsfonts}       
\usepackage{nicefrac}       
\usepackage{microtype}      
\usepackage{xcolor}         
\usepackage{amsmath}

\title{CCF: Cross Correcting Framework for Pedestrian Trajectory Prediction}

\author{
  Pranav Singh Chib \\
CSE Department\\
  IIT Roorkee, India \\
  \texttt{pranavs\_chib\@cs.iitr.ac.in} \\
  \And
  Pravendra Singh \\
  CSE Department \\
 IIT Roorkee, India \\
  \texttt{pravendra.singh@cs.iitr.ac.in} \\
}

\begin{document}

\maketitle

\begin{abstract}

Accurately predicting future pedestrian trajectories is crucial across various domains. Due to the uncertainty in future pedestrian trajectories, it is important to learn complex spatio-temporal representations in multi-agent scenarios. To address this, we propose a novel Cross-Correction Framework (CCF) to learn spatio-temporal representations of pedestrian trajectories better. Our framework consists of two trajectory prediction models, known as subnets, which share the same architecture and are trained with both cross-correction loss and trajectory prediction loss. Cross-correction leverages the learning from both subnets and enables them to refine their underlying representations of trajectories through a mutual correction mechanism. Specifically, we use the cross-correction loss to learn how to correct each other through an inter-subnet interaction. To induce diverse learning among the subnets, we use the transformed observed trajectories produced by a neural network as input to one subnet and the original observed trajectories as input to the other subnet. We utilize transformer-based encoder-decoder architecture for each subnet to capture motion and social interaction among pedestrians. The encoder of the transformer captures motion patterns in trajectories, while the decoder focuses on pedestrian interactions with neighbors. Each subnet performs the primary task of predicting future trajectories (a regression task) along with the secondary task of classifying the predicted trajectories (a classification task). Extensive experiments on real-world benchmark datasets such as ETH-UCY and SDD demonstrate the efficacy of our proposed framework, CCF,  in precisely predicting pedestrian future trajectories. We also conducted several ablation experiments to demonstrate the effectiveness of various modules and loss functions used in our approach.

\end{abstract}

\section{Introduction}

Pedestrian trajectory prediction aims to forecast pedestrians’ future paths based on their past observed trajectories. In recent years, trajectory prediction has been applied to various fields, including autonomous driving, robotics, surveillance, and human behavior analysis. Accurate trajectory forecasting is crucial for the safety and advancement of these applications. Pedestrian dynamics can be uncertain due to multiple possible future trajectories. Additionally, the temporal motion of a pedestrian depends on their intentions and spatial interactions \cite{sophie19,kosaraju2019social,lv2023ssagcn,sun2022interaction} with neighboring pedestrians. Therefore, it is important to capture the spatio-temporal representation \cite{tsao2022social,liang2020learning,bae2022gpgraph,wu2023multi,girgis2022latent} of pedestrians for accurate trajectory prediction.

Most of the trajectory prediction methods \cite{shi2021sgcn, mohamed2020social, sekhon2021scan, gupta2018social} use distinct models to learn temporal motion, subsequently giving the trajectories into separate temporal models to extract temporal features. These temporal models often use sequential networks, including Recurrent Neural Networks (RNNs) \cite{cao2018brits,chiara2022goal}, Temporal Convolutional Networks (TCNs) \cite{bai2018empirical,bae2021disentangled,mohamed2020social,mendieta2021carpe}, Self-attention mechanisms \cite{vaswani2017attention,yang2023social,duan2022complementary,kosaraju2019social,shafiee2021introvert}, and Transformers \cite{vaswani2017attention,tsao2022social,yuan2021agentformer,shi2023trajectory,girgis2022latent}, to capture temporal motion. Additionally, some works have utilized Knowledge Distillation (KD) \cite{9879765,das2023distilling} to transfer spatio-temporal knowledge to the student model, consequently improving student performance. In STT \cite{9879765}, the teacher-student paradigm is used to transfer knowledge from the teacher model to the student model, enabling the student network to make predictions with fewer input observations. Similarly, Di-Long \cite{das2023distilling} distills knowledge from an expert large model trained on short-term predictions to guide the student in long-term trajectory prediction. Although these approaches have shown promising results, they are limited by the performance of the teacher model \cite{mirzadeh2020improved,cho2019efficacy} and whether the distilled model can capture all the intricate details and knowledge from the teacher model. In contrast, we propose a cross-correcting framework for trajectory prediction that allows each subnetwork to learn to correct each other through inter-subnet interaction without the need for a teacher model.

In this work, we propose a novel \textbf{C}ross \textbf{C}orrecting \textbf{F}ramework (CCF) for trajectory prediction to better learn spatio-temporal representations of pedestrian trajectories. Specifically, our framework consists of two trajectory prediction models, known as subnets, which share the same architecture and are trained in parallel to capture pedestrian motion dynamics and social interactions (see Fig.~\ref{supply_fig_crosscorr}). To induce diversity among the subnets, we utilize a DNet network that transforms the observed trajectories into diversified versions of the past observed trajectories (see Sec.~\ref{sec:diversity}). One subnet receives the original past observed trajectories, while the other subnet receives the transformed trajectories to promote diverse learning between the two subnets. Using cross-correction loss, we leverage the learning from both subnets through mutual correction. This inter-subnet interaction allows them to learn how to correct each other, enabling them to refine their representations of underlying trajectories. Each subnet is a transformer-based encoder-decoder architecture; the encoder captures motion patterns in trajectories, while the decoder focuses on pedestrian interactions with neighbors. Additionally, each subnet performs the primary task of predicting future trajectories (a regression task) along with the secondary task of classifying the predicted trajectories (a classification task), where the secondary task aids the primary task (see Sec.\ref{sec:componets}). We cluster future trajectories of pedestrians from the training data to capture general motion patterns. Then, we use each cluster mean as a future trajectory class to formulate it as a classification task. Each subnet also predicts the class probabilities of the predicted future trajectories (see Fig.\ref{supply_fig_scean}) to learn the general motion patterns in future trajectories. Experimental results demonstrate that our proposed CCF results in accurate future trajectory predictions on real-world pedestrian trajectory prediction ETH-UCY and SDD benchmark datasets. We also conducted several ablation experiments to demonstrate the effectiveness of the various modules and loss functions used in our approach.

\section{Related Work}

\subsection{Trajectory prediction}  
The trajectory prediction model aims to predict future trajectories using past observed trajectories. Prior research on trajectory prediction mostly focused on deterministic approaches, such as force models \cite{helbing1995social}, sequential models \cite{salzmann2020trajectron++}, and frequency analysis \cite{wong2022view} methods. These models \cite{helbing1995social} use parameters like acceleration and target velocity to model the trajectories followed by objects or agents. These trajectory sequences can be modeled using sequence-to-sequence techniques due to the potential changes in future states from the current ones. Significant advancements have been made in sequence prediction problems with the use of Recurrent Neural Networks (RNNs) \cite{alahi2016social, huang2019stgat} and Long Short-Term Memory networks (LSTM) \cite{hochreiter1997long}. These architectures have been used in learning the temporal patterns of pedestrian movements. Additionally, spatio-temporal networks, which can represent structured sequence data, have been developed using LSTM networks \cite{ivanovic2019trajectron, vemula2018social}. However, it is important to note that RNN-based models can sometimes face challenges such as vanishing or exploding gradients. A variety of future trajectories can be possible when predicting the trajectory of an agent. Stochastic prediction models are also used to capture such uncertainty in predicting the future trajectory, such as conditional variational autoencoders (CVAEs) \cite{Xu_2022_CVPR,lee2022muse,xu2022dynamic,mangalam2020not} and generative adversarial networks (GANs) \cite{sophie19, hu2020collaborative,gupta2018social,kosaraju2019social,gupta2018social}. Recently, diffusion models \cite{mao2023leapfrog} have also been used in predicting future trajectories. However, diffusion methods involve high inference times because they involve multiple denoising steps. LED \cite{mao2023leapfrog} mitigates this issue by incorporating a trainable leapfrog initializer. Transformers are also used because they capture long-term temporal dependencies due to the attention mechanism \cite{girgis2022latent,yuan2021agentformer,zhou2023query,yu2020spatio}. Some techniques \cite{mohamed2020social,mendieta2021carpe,bae2021disentangled} use a convolution neural network to encode the scene image or traffic map in order to include the environmental information. Additionally, in order to represent agent interactions, graph-based techniques \cite{lv2023ssagcn,sekhon2021scan,pmlr-v80-kipf18a} have also been used. Some works \cite{lv2023ssagcn,yu2020spatio,mohamed2020social,yu2020spatio} used a graph structure to capture social and scene interactions. Graph neural networks with attention mechanisms \cite{wang2023trajectory,zhou2023static} are also applied to capture various types of interactions, such as single, pairwise, or group interactions, to predict future trajectories. Trajectory prediction has also been the subject of numerous other approaches, including long-tail trajectory prediction \cite{wang2023ganet}, endpoint-conditioned trajectory prediction \cite{bae2023set}, and others.

\subsection{Spatio-Temporal Learning}

Various architectures have been developed for trajectory prediction models to model spatio-temporal patterns, such as \cite{alahi2016social,huang2019stgat,Hu2022vehicle,li2023multi}. Social-LSTM \cite{alahi2016social} uses LSTM to model spatio-temporal interactions. Similarly, models like Trajectron \cite{ivanovic2019trajectron}, and Social Attention \cite{vemula2018social} use LSTM to create spatio-temporal graphs that represent structured sequence trajectories. RNN-based models \cite{salzmann2020trajectron++,park2020diverse} are also used in learning Spatio-temporal representations. Additionally, some models, such as SGCN \cite{shi2021sgcn}, employ Temporal Convolutional Networks (TCNs) \cite{lv2023ssagcn} to learn spatio-temporal representations. Transformers have been employed to model temporal dependencies by several works \cite{girgis2022latent,yuan2021agentformer,zhou2023query}. Yuan et al. \cite{yuan2021agentformer} characterize the social and temporal representation of agents by aggregating trajectory attributes across agents and time to represent multi-agent trajectories. Zhong et al. \cite{zhong2023visual} utilize a spatio-temporal Transformer that integrates visual information to describe pedestrian motion. Wu et al. \cite{wu2023multi} focus on learning spatio-temporal correlation features using a sparse attention gate to capture spatio-temporal information. In our approach, we use cross-correction loss to facilitate mutual correction through inter-subnet interaction, enhancing the model's ability to better capture the spatio-temporal characteristics of pedestrian trajectories. To promote diverse learning among the subnets, we feed transformed observed trajectories produced by a neural network to one subnet while the other subnet receives the original observed trajectories as input.


\section{Method}

\subsection{Formulation}

Pedestrian trajectory prediction involves analyzing a pedestrian's observed past trajectories as well as those of its neighbors to predict the pedestrian's future trajectory. Formally, the past observed trajectory with $T_{\text{ob}}$ timestamps is represented by $\bm{X}_i = \{ (x_i^t, y_i^t)  \mid  t \in [1, \dots, T_{\text{ob}}] \}$, where $(x_i^t, y_i^t)$ denotes the 2D spatial coordinates of pedestrian $i$ at time $t$. Similarly, the future ground truth trajectory for the time duration $T_{\text{pred}}$ to be predicted is represented as $\bm{Y}_i = \{ (x_i^t, y_i^t) \mid t \in [T_{\text{ob}} + 1, \ldots, T_{\text{pred}}] \}$. The scene contains $N$ pedestrians, and $i \in N$ represents the pedestrian index. The aim of the trajectory prediction model is to generate future trajectories $\bm{\hat{Y}_i}$ that closely match with the ground truth trajectories $\bm{Y}_i$. Given the indeterminacy of future movements, most works \cite{mao2023leapfrog, Xu_2022_CVPR,bae2023set} predict more than one trajectory to capture this uncertainty. We present formulations for a single pedestrian and a batch size of one throughout the paper, which can be generalized for $N$ pedestrians and $B$ batches.

\subsection{CCF: Cross Correcting Framewok}

\begin{figure*}[!t]
    \centering
    \includegraphics[scale=.20]{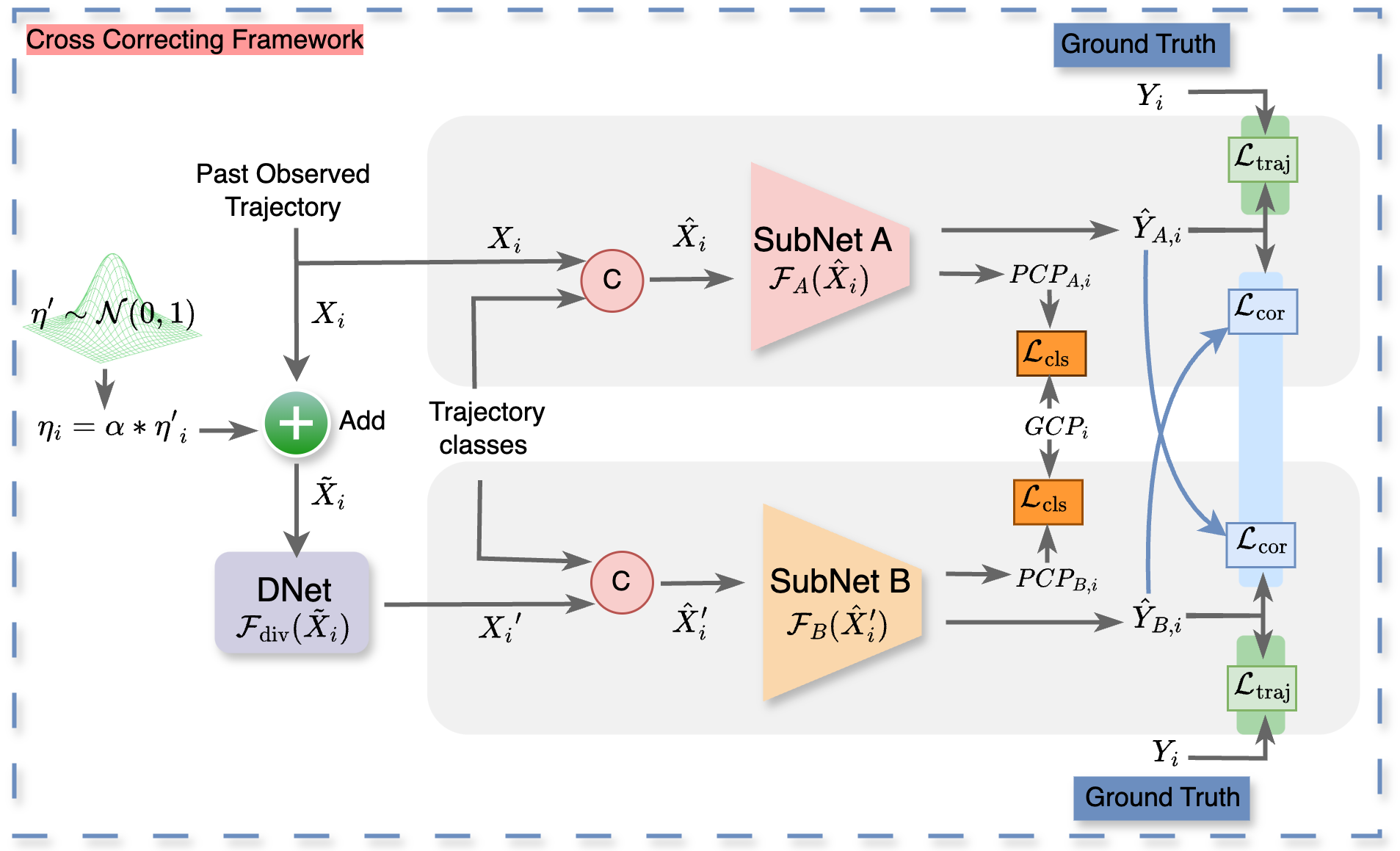}

\caption{
The illustration of the CCF framework consists of two transformer-based trajectory prediction networks (SubNet A and SubNet B). SubNet A receives the original observed trajectory ($X_i$), while SubNet B is provided with a diverse version (${X_i}^{\prime}$) of $X_i$. First, Gaussian noise is added to $X_i$, and then it is given as an input to DNet to get a diversified version (${X_i}^{\prime}$). The original observed trajectory $X_i$ and the transformed trajectory ${X_i}^{\prime}$ are separately concatenated (symbol C) with trajectory classes given as inputs to SubNet A and SubNet B. The cross-correction mechanism is used for inter-subnet interactions. Here, $PCP_{A,i}$ and $PCP_{B,i}$ are the predicted class probabilities by SubNet A and SubNet B, respectively, and $GCP_{i}$ is the ground truth class probabilities. $\mathcal{L}_{\text{traj}}$ is the trajectory prediction loss, and $\mathcal{L}_{\text{cor}}$ is the cross-correction loss.}
\label{supply_fig_crosscorr}    
\end{figure*}

Our proposed CCF comprises two subnets (abbreviation for two models), denoted as subnet A ($\mathcal{F}_A(\cdot)$) and subnet B ($\mathcal{F}_B(\cdot)$), both exhibiting diverse learning despite having the same architecture (Sec. \ref{sec:subnets}). Both subnets are parallelly trained to predict future trajectory (Fig. \ref{supply_fig_crosscorr}). To diversify the learning between the two subnets, we utilize diverse versions of the observed past trajectories. One subnet is trained with the original past observed trajectory ($X_i$), while the other subnet is trained with a transformed version of the past observed trajectory (${X_i}^{\prime}$). We diversify the observed trajectory using the DNet ($\mathcal{F}_{div}(\cdot)$) (Sec. \ref{sec:Dnet}), which transforms the past observed trajectory into a diverse version of it. We cluster future trajectories of pedestrians from the training data to generate trajectory classes and then use each cluster mean as a future trajectory class. Thus, given the observed trajectory concatenated with trajectory classes, subnet $\mathcal{F}_A$ predicts the trajectory $\hat{Y}_{A,i}$. Similarly, subnet $\mathcal{F}_B$ predicts $\hat{Y}_{B,i}$ given the diverse version of observed trajectory concatenated with trajectory classes, as shown in Eqs. \ref{eq_1}, \ref{eq_2} and Fig. \ref{supply_fig_crosscorr}.

\begin{align} 
    \hat{Y}_{A,i} = \mathcal{F}_A(\hat{X_i}) \label{eq_1}\\ 
    \hat{Y}_{B,i} = \mathcal{F}_B({\hat{X}_i}\sp{\prime}) \label{eq_2}
\end{align}

\subsection{DNet Network} \label{sec:Dnet}
Given the observed past trajectory $X_{i}$, we sample Gaussian noise and add it to $X_{i}$ to generate $\tilde{X}_{i}$ as shown in Fig. \ref{supply_fig_crosscorr}. Specifically, ${\eta^{\prime}}_{i}$ is the noise sampled from a standard normal distribution $\mathcal{N}(0, 1)$. We control the magnitude of this noise using a parameter $\alpha$, referred to as the noise factor, to obtain the final additive noise $\eta_{i} = \alpha*{\eta^{\prime}}_{i} \label{eq_noise_addd}$. We add additive noise $\eta_{i}$ into the past observed trajectory to create $\tilde{X}_{i}$, where $\tilde{X}_{i} = {X}_{i}+ {\eta}_{i}$. Subsequently, $\tilde{X}_{i}$ is passed as an input to the DNet network, which generates a diverse version (${X_i}^{\prime}$) of $X_{i}$ as shown in Eq. \ref{eq:dnet}.

\begin{align} \label{eq:dnet} 
    {X_i}^{\prime} = \mathcal{F}_{\text{div}}(\tilde{X}_{i})
\end{align}

Where, DNet network $\mathcal{F}_{\text{div}}(.)$ is a multi-layer perceptron with two hidden layers.

The training loss $\mathcal{L}_{\text{div}}$ of DNet is given in Eq. \ref{eq:dnetloss}. We show empirically that the output of the DNet network does not exactly replicate the original past observed trajectory but instead generates a diverse version of it (Sec. \ref{sec:diversity}). We also experimented with directly adding noise $\eta_{i}$ to create a diverse version of past observed trajectory but obtained suboptimal results because $\eta_{i}$ randomly relocated coordinates of past observed trajectory (ref Table \ref{tab:diverseNet}).

\begin{equation} \label{eq:dnetloss}
\mathcal{L}_{\text{div}} =  \mathcal{L}_{\text{Huber}}(X_i, {X_i}^{\prime})
\end{equation}

\begin{figure}[t]
    \centering
    \includegraphics[scale=.26]{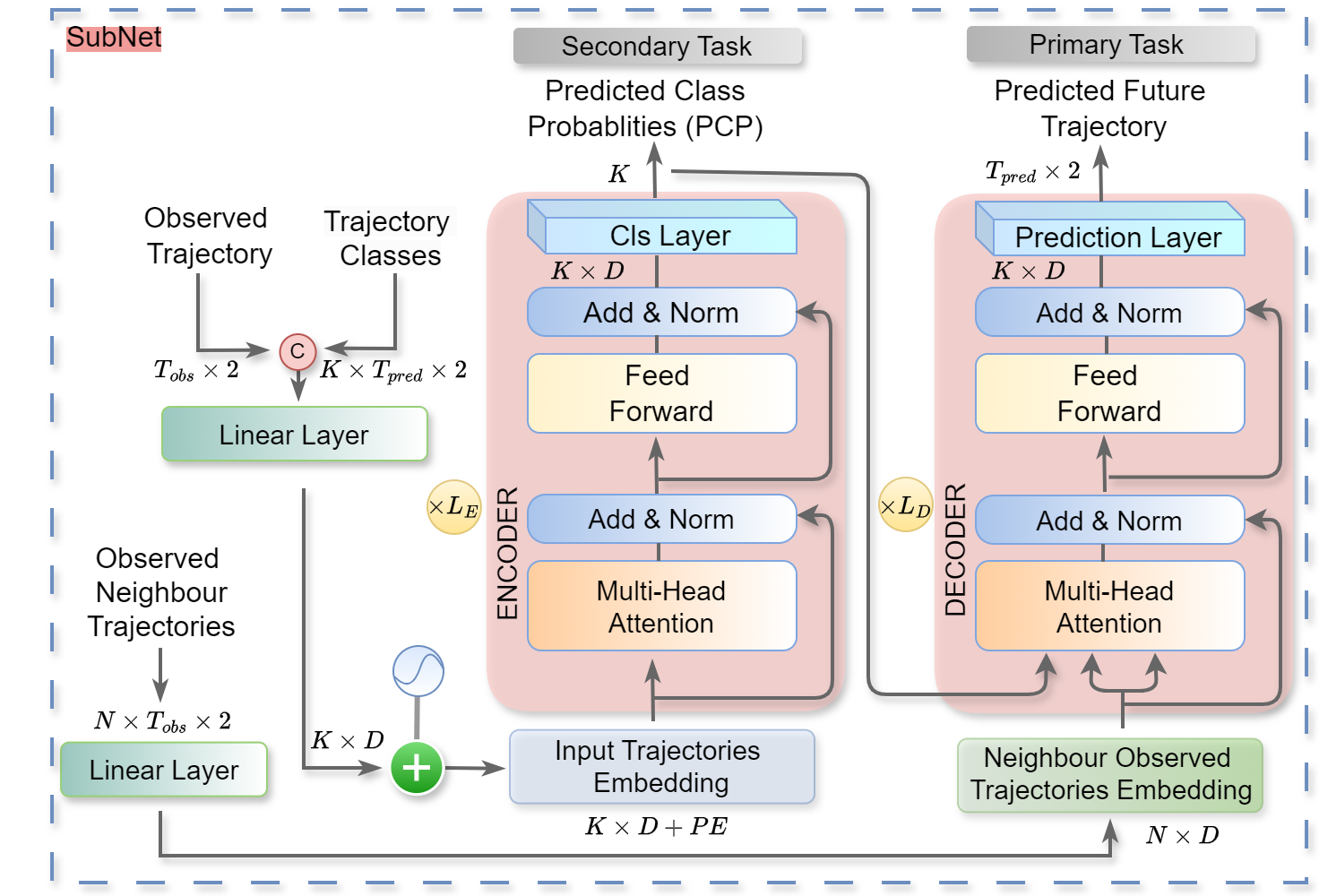}
 \caption{
Illustration of the transformer-based subnet. The observed trajectory is concatenated (symbol C) with trajectory classes and given as an input to a linear layer. The output of the linear layer is added with position encoding to get input embedding. This input embedding is passed as an input to the encoder of the transformer. The output of the encoder is the predicted class probabilities. Next, the output of the encoder, along with the neighboring observed trajectories embedding, is fed to the decoder to capture the social interaction of the pedestrian and their neighbors. Finally, the decoder outputs the predicted future trajectory. Here, K denotes the number of trajectory classes, D denotes the embedding size, PE refers to the position encoding, and N indicates the number of neighbors. $L_E$ and $L_D$ represent the number of layers in the encoder and decoder, respectively.}
    \label{supply_fig_scean}
\end{figure}

\subsection{SubNet - Transformer Architecture} \label{sec:subnets}

We first concatenate the past observed trajectory $\in \mathbb{R}^{ T_{\text{ob}} \times 2}$ and the trajectory classes $\in \mathbb{R}^{K \times T_{pred} \times 2}$ and then apply linear transformation by a linear layer to create an encoder-compatible dimensional embedding $\in \mathbb{R}^{ K \times D}$ followed by addition of a sinusoidal position embedding ($PE$) to capture the temporal dependencies present in the underlying trajectory. Where $K$ is the number of trajectory classes, and $D$ is the embedding size. Similarly, for the decoder, the neighbor embedding $\in \mathbb{R}^{N \times D}$ is generated by applying a linear transformation to past observed trajectories of neighbors  $\in \mathbb{R}^{N \times T_{obs} \times 2}$. Here, $N$ represents the number of neighbors. We use standard transformer architecture in our approach. Each layer of the encoder consists of multi-head self-attention and a feed-forward network. The encoder output is fed to the linear layer head (Cls layer) to predict the class probabilities (secondary task) as shown in Fig. \ref{supply_fig_scean}. The decoder takes encoder output and neighbors embedding $\in \mathbb{R}^{N \times D}$ as input to predict the future trajectory of pedestrian. The decoder captures social interactions among pedestrians and their neighbors. The output of the decoder is the predicted trajectory $\in \mathbb{R}^{ T_{\text{pred}}\times 2}$ of the pedestrian (primary task). The trajectory prediction loss function of the subnet is given below in Eq. \ref{eq:tp_losss}, which is a Huber loss function.

\begin{equation} \label{eq:tp_losss}
\mathcal{L}_{\text{traj}} =  \mathcal{L}_{\text{Huber}}(Y_i, \hat{Y}_i)
\end{equation}

\textbf{Secondary Task.}

The secondary task of each subnet is to predict class probabilities ($PCP_i$). We cluster future trajectories of pedestrians from the training data to capture general motion patterns. Each cluster mean is then used as a future trajectory class to formulate the classification task. We utilize k-mean clustering  to create $K$ classes $\{ c_1, c_2, \dots, c_{K} \}$. The ground truth class probabilities ($GCP_i$) are calculated by the negative $l_2$ distance between the ground truth trajectory $Y_i$ and the trajectory classes, followed by softmax. Specifically, $ GCP_i = softmax \{-\| {Y_i} - c_{j} \|_2^2 ~|~j \in \{ 1,2, \dots, K \}\} $ where, K is total classes and $Y_i$ is the ground truth future trajectory of the $i^{th}$ pedestrian.
We use cross-entropy loss function $\mathcal{L}_{CC}$ for the classification task as shown below.

\begin{equation}
\mathcal{L}_{\text{cls}} = \mathcal{L}_{\text{CC}}(GCP_i, P CP_i)
\end{equation}

The final loss function for each subnet is given below.

\begin{equation}
    \mathcal{L}_{\text{subnet}} = \mathcal{L}_{\text{traj}} + \mathcal{L}_{\text{cls}}
     \label{eq:sub_final}
\end{equation}

Where, $\mathcal{L}_{\text{traj}}$ is the trajectory prediction loss and $
\mathcal{L}_{\text{cls}}$ is the trajectory classification loss.

\subsection{Cross Correction Loss}
It is important to facilitate cross-correction between subnets to enable them to correct each other through inter-subnet interaction. Effective cross-correction enhances the capability of subnets to accurately predict future trajectories. We utilize the Huber loss function for cross-correction loss, as given below.

\begin{equation}
    \mathcal{L}_{\text{cor,A}} =
    \mathcal{L}_{\text{Huber, B$ \rightarrow $A}} (\hat{Y}_{B,i}, \hat{Y}_{A,i}) 
    \label{eq:8}
\end{equation} 

\begin{equation}
    \mathcal{L}_{\text{cor,B}} = \mathcal{L}_{\text{Huber, A$ \rightarrow $B}} (\hat{Y}_{A,i}, \hat{Y}_{B,i})
    \label{eq:9}
\end{equation}

Eq.~\ref{eq:8} shows the cross correction loss for subnet A ($\mathcal{L}_{\text{cor,A}}$), where subnet B is correcting subnet A and $\hat{Y}_{B,i}$ is the ground truth for $\mathcal{L}_{\text{cor,A}}$ loss. Similarly, subnet A is correcting subnet B using $\mathcal{L}_{\text{cor,B}}$.

\subsection{Training and Evaluation}
We can train the trajectory prediction model using our proposed cross-correcting framework (CCF) with the following loss function:

\begin{equation}
    \mathcal{L}_{\text{total}} = \mathcal{L}_{\text{div}} + \mathcal{L}_{\text{subnet,A}} + \mathcal{L}_{\text{subnet,B}} + \lambda(\mathcal{L}_{\text{cor,A}} + \mathcal{L}_{\text{cor,B}})
\end{equation}

Here, $\mathcal{L}_{\text{div}}$ is the DNet loss (Eq. \ref{eq:dnetloss}), $\mathcal{L}_{\text{subnet,A}}$ (Eq. \ref{eq:sub_final}) represents the loss for training subnet A , $\mathcal{L}_{\text{subnet,B}}$ represents the loss for training subnet B,  
$\mathcal{L}_{\text{cor,A}}$ represents the cross correction loss for subnet A (Eq.~\ref{eq:8}), and $\mathcal{L}_{\text{cor,B}}$ represents the cross correction loss for subnet B (Eq.~\ref{eq:8}). The $\lambda$ is a hyperparameter that is used to weigh the contribution of the cross-correction loss in the total loss. 

For evaluation during testing, we use only one subnet. Empirically, we have found that both subnets perform equally well. Therefore, we use subnet A for reporting results in the experimental section.

\begin{table}[t]
\caption{Comparison of our approach (CCF) with other approaches on ETH, HOTEL, UNIV, ZARA1, and ZARA2 in terms of ADE/FDE ($\downarrow$ values are better). The best results are highlighted in \textbf{bold}, and the second-best results are shown in \underline{underlined}.}
\label{eth_main}
\resizebox{\columnwidth}{!}{%
\begin{tabular}{@{}c|c|ccccc|c@{}}
\toprule  \toprule
Model &
  Venue &
  ETH &
  HOTEL &
  UNIV &
  ZARA1 &
  ZARA2 &
  AVG \\ \midrule 
NMMP &
  CVPR 2020 &
  \multicolumn{1}{c|}{0.62/1.08} &
  \multicolumn{1}{c|}{0.33/0.63} &
  \multicolumn{1}{c|}{0.52/1.11} &
  \multicolumn{1}{c|}{0.32/0.66} &
  0.29/0.61 &
  0.41/0.82 \\ \midrule
Social-STGCNN &
  CVPR 2020 &
  \multicolumn{1}{c|}{0.64/1.11} &
  \multicolumn{1}{c|}{0.49/0.85} &
  \multicolumn{1}{c|}{0.44/0.79} &
  \multicolumn{1}{c|}{0.34/0.53} &
  0.30/0.48 &
  0.44/0.75 \\ \midrule
PecNet &
  ECCV 2020 &
  \multicolumn{1}{c|}{0.54/0.87} &
  \multicolumn{1}{c|}{0.18/0.24} &
  \multicolumn{1}{c|}{0.35/0.60} &
  \multicolumn{1}{c|}{0.22/0.39} &
  0.17/0.30 &
  0.29/0.48 \\ \midrule
Trajectron++ &
  ECCV 2020 &
  \multicolumn{1}{c|}{0.61/1.03} &
  \multicolumn{1}{c|}{0.20/0.28} &
  \multicolumn{1}{c|}{0.30/0.55} &
  \multicolumn{1}{c|}{0.24/0.41} &
  0.18/0.32 &
  0.31/0.52 \\ \midrule
SGCN &
  CVPR 2021 &
  \multicolumn{1}{c|}{0.52/1.03} &
  \multicolumn{1}{c|}{0.32/0.55} &
  \multicolumn{1}{c|}{0.37/0.70} &
  \multicolumn{1}{c|}{0.29/0.53} &
  0.25/0.45 &
  0.37/0.65 \\ \midrule
STGAT &
  AAAI 2021 &
  \multicolumn{1}{c|}{0.56/1.10} &
  \multicolumn{1}{c|}{0.27/0.50} &
  \multicolumn{1}{c|}{0.32/0.66} &
  \multicolumn{1}{c|}{0.21/0.42} &
  0.20/0.40 &
  0.31/0.62 \\ \midrule
CARPE &
  AAAI 2021 &
  \multicolumn{1}{c|}{0.80/1.40} &
  \multicolumn{1}{c|}{0.52/1.00} &
  \multicolumn{1}{c|}{0.61/1.23} &
  \multicolumn{1}{c|}{0.42/0.84} &
  0.34/0.74 &
  0.46/0.89 \\ \midrule
AgentForme &
  ICCV 2021 &
  \multicolumn{1}{c|}{0.45/0.75} &
  \multicolumn{1}{c|}{0.14/0.22} &
  \multicolumn{1}{c|}{0.25/0.45} &
  \multicolumn{1}{c|}{0.18/0.30} &
  0.14/0.24 &
  0.23/0.39 \\ \midrule
GroupNet &
  CVPR 2022 &
  \multicolumn{1}{c|}{0.46/0.73} &
  \multicolumn{1}{c|}{0.15/0.25} &
  \multicolumn{1}{c|}{0.26/0.49} &
  \multicolumn{1}{c|}{0.21/0.39} &
  0.17/0.33 &
  0.25/0.44 \\ \midrule
GP-Graph &
  ECCV 2022 &
  \multicolumn{1}{c|}{0.43/0.63} &
  \multicolumn{1}{c|}{0.18/0.30} &
  \multicolumn{1}{c|}{0.24/0.42} &
  \multicolumn{1}{c|}{0.17/0.31} &
  0.15/0.29 &
  0.23/0.39 \\ \midrule
STT &
  CVPR 2022 &
  \multicolumn{1}{c|}{0.54/1.10} &
  \multicolumn{1}{c|}{0.24/0.46} &
  \multicolumn{1}{c|}{0.57/1.15} &
  \multicolumn{1}{c|}{0.45/0.94} &
  0.36/0.77 &
  0.43/0.88 \\ \midrule
Social-Implicit &
  ECCV 2022 &
  \multicolumn{1}{c|}{0.66/1.44} &
  \multicolumn{1}{c|}{0.20/0.36} &
  \multicolumn{1}{c|}{0.31/0.60} &
  \multicolumn{1}{c|}{0.25/0.50} &
  0.22/0.43 &
  0.33/0.67 \\ \midrule
BCDiff &
  NIPS 2023 &
  \multicolumn{1}{c|}{0.53/0.91} &
  \multicolumn{1}{c|}{0.17/0.27} &
  \multicolumn{1}{c|}{0.24/0.40} &
  \multicolumn{1}{c|}{0.21/0.37} &
  0.16/0.26 &
  0.26/0.44 \\ \midrule
Graph-TERN &
  AAAI 2023 &
  \multicolumn{1}{c|}{0.42/0.58} &
  \multicolumn{1}{c|}{0.14/0.23} &
  \multicolumn{1}{c|}{0.26/0.45} &
  \multicolumn{1}{c|}{0.21/0.37} &
  0.17/0.29 &
  0.24/0.38 \\ \midrule
FlowChain &
  ICCV 2023 &
  \multicolumn{1}{c|}{0.55/0.99} &
  \multicolumn{1}{c|}{0.20/0.35} &
  \multicolumn{1}{c|}{0.29/0.54} &
  \multicolumn{1}{c|}{0.22/0.40} &
  0.20/0.34 &
  0.29/0.52 \\ \midrule
EigenTrajectory &
  ICCV 2023 &
  \multicolumn{1}{c|}{\textbf{0.36} /\underline{0.56}} &
  \multicolumn{1}{c|}{0.14/0.22} &
  \multicolumn{1}{c|}{\underline{0.24}/0.43} &
  \multicolumn{1}{c|}{0.21/0.39} &
  0.16/0.29 &
  0.23/0.38 \\ \midrule
SMEMO &
  TPAMI 2024 &
  \multicolumn{1}{c|}{0.39/0.59} &
  \multicolumn{1}{c|}{\underline{0.14}/\underline{0.20}} &
  \multicolumn{1}{c|}{0.23/\textbf{0.41}} &
  \multicolumn{1}{c|}{\underline{0.19}/\textbf{0.32}} &
  \underline{0.15}/\underline{0.26} &
  \underline{0.22}/\underline{0.35} \\ \midrule
CCF (Our) &
  - &
  \multicolumn{1}{c|}{\underline{0.38}/\textbf{0.55}} &
  \multicolumn{1}{c|}{\textbf{0.11}/\textbf{0.17}} &
  \multicolumn{1}{c|}{\textbf{0.23}/\underline{0.42}} &
  \multicolumn{1}{c|}{\textbf{0.18}/\underline{0.34}} &
  \textbf{0.14}/\textbf{0.25} &
  \textbf{0.20}/\textbf{0.34} \\ \bottomrule
\end{tabular}%
}
\end{table}

\section{Experiments}

\subsection{Experimental Settings}

\textbf{Datasets}. We evaluate our approach using two popular pedestrian trajectory prediction benchmarks: ETH-UCY \cite{lerner2007crowds,pellegrini2009you} and the Stanford Drone Dataset (SDD) \cite{robicquet2016learning}. ETH-UCY consists of five scenarios, totaling 1536 human trajectories with diverse behaviors such as walking, crossing, grouping, and following. The SDD dataset comprises 20 unique scenes captured from a top-down drone view, containing a variety of agents navigating within a university campus, including pedestrians, bikers, skateboarders, cars, buses, and golf carts, totaling 11,000 agents with 5,232 trajectories. The trajectories in ETH-UCY are labeled in meters, whereas in SDD, trajectory positions are recorded in pixel coordinates. Following prior works \cite{mangalam2020not,hu2020collaborative}, our evaluation protocol includes an observation length of 8 timesteps (3.2s) and a prediction horizon of 12 timesteps (4.8s) for both datasets. We use a leave-one-out approach for the training and evaluation. 

\textbf{Evaluation Metrics.} 
We use two evaluation metrics to assess the proposed method: Average Displacement Error (ADE) and Final Displacement Error (FDE). ADE measures the average $l_2$ distance between the predicted trajectory and the actual trajectory across all predicted time steps, while FDE measures the $l_2$ distance between the predicted final destination and the ground truth final destination at the end of the prediction time steps. Consistent with prior works \cite{alahi2016social,gupta2018social}, we generate 20 trajectories and select the best based on minimum ADE (minADE$_{20}$) and minimum FDE (minFDE$_{20}$).

\textbf{Implementation Details.} We conducted the experiments using Python 3.8.13 and PyTorch version 1.13.1+cu117. The training phase took place on an NVIDIA RTX A5000 GPU with an AMD EPYC 7543 CPU. Training time for a single epoch on the ETH is 20.15 seconds using our approach CCF. We observed a slight increase (approximately 1.4 times) in training time compared to without using CCF for a single epoch on the ETH.  Nevertheless, our approach does not impact test time (test time is 0.463 seconds on complete test data of ETH) since we only use one model during testing. In the subnet architecture, we use a single encoder layer for the ETH and HOTEL datasets and two encoder layers for UNIV, ZARA1, ZARA2, and SDD datasets. Each dataset uses one decoder layer. We utilize a mask in the decoder to prevent attention from focusing on invalid agents. We introduce Gaussian noise with a zero mean and a standard deviation of 1. The noise magnitude is controlled by the noise factor $\alpha$, which is experimentally set to 0.1 for ETH and HOTEL, and 0.01 for UNIV, ZARA1, ZARA2, and SDD. The contribution of the cross-correction loss is set to 0.1 (refer to Section \ref{sec:abl}).

\begin{table}[!t]
\caption{Comparison of our approach (CCF) with other approaches on SDD dataset in terms of ADE/FDE ($\downarrow$ values are better). The best results are highlighted in \textbf{bold}, and the second-best results are shown in \underline{underlined}.}
\label{tab:sddmain}
\resizebox{\columnwidth}{!}{%
\begin{tabular}{@{}c|ccccccc@{}}
\toprule  \bottomrule
Model &
  SGCN &
  LBEBM &
  PCCSNet &
  GroupNet &
  MemoNet &
  CAGN &
  SIT \\ \midrule 
Venue &
  \multicolumn{1}{c|}{CVPR 2021} &
  \multicolumn{1}{c|}{CVPR 2021} &
  \multicolumn{1}{c|}{ICCV 2021} &
  \multicolumn{1}{c|}{CVPR 2022} &
  \multicolumn{1}{c|}{CVPR 2022} &
  \multicolumn{1}{c|}{AAAI 2022} &
  AAAI 2022 \\ \midrule
ADE &
  \multicolumn{1}{c|}{11.67} &
  \multicolumn{1}{c|}{9.03} &
  \multicolumn{1}{c|}{8.62} &
  \multicolumn{1}{c|}{0.46/0.73} &
  \multicolumn{1}{c|}{1.00/2.08} &
  \multicolumn{1}{c|}{9.42} &
  9.13 \\ \midrule
FDE &
  \multicolumn{1}{c|}{19.10} &
  \multicolumn{1}{c|}{15.97} &
  \multicolumn{1}{c|}{16.16} &
  \multicolumn{1}{c|}{0.15/0.25} &
  \multicolumn{1}{c|}{0.35/0.67} &
  \multicolumn{1}{c|}{15.93} &
  15.42 \\ \midrule \bottomrule
Model & 
  \multicolumn{1}{c|}{MID} &
  \multicolumn{1}{c|}{SocialVAE} &
  \multicolumn{1}{c|}{Graph-TERN} &
  \multicolumn{1}{c|}{BCDiff} &
  \multicolumn{1}{c|}{MRL} &
  \multicolumn{1}{c|}{SMEMO} &
  CCF (Our) \\ \midrule
Venue &
  \multicolumn{1}{c|}{CVPR 2022} &
  \multicolumn{1}{c|}{ECCV 2022} &
  \multicolumn{1}{c|}{AAAI 2023} &
  \multicolumn{1}{c|}{NeuIPS 2023} &
  \multicolumn{1}{c|}{AAAI 2023} &
  \multicolumn{1}{c|}{TPAMI 2024} &
  - \\ \midrule
ADE &
  \multicolumn{1}{c|}{9.73} &
  \multicolumn{1}{c|}{8.88} &
  \multicolumn{1}{c|}{8.43} &
  \multicolumn{1}{c|}{9.05} &
  \multicolumn{1}{c|}{8.22} &
  \multicolumn{1}{c|}{{\underline{ 8.11}}} &
  \textbf{7.82} \\ \midrule
FDE &
  \multicolumn{1}{c|}{15.32} &
  \multicolumn{1}{c|}{14.81} &
  \multicolumn{1}{c|}{14.26} &
  \multicolumn{1}{c|}{14.86} &
  \multicolumn{1}{c|}{13.39} &
  \multicolumn{1}{c|}{{\underline{13.06}}} &
  \textbf{12.77} \\ \bottomrule
\end{tabular}%
}
\end{table}

\begin{figure}[t]
    \centering
    \includegraphics[scale=.14]{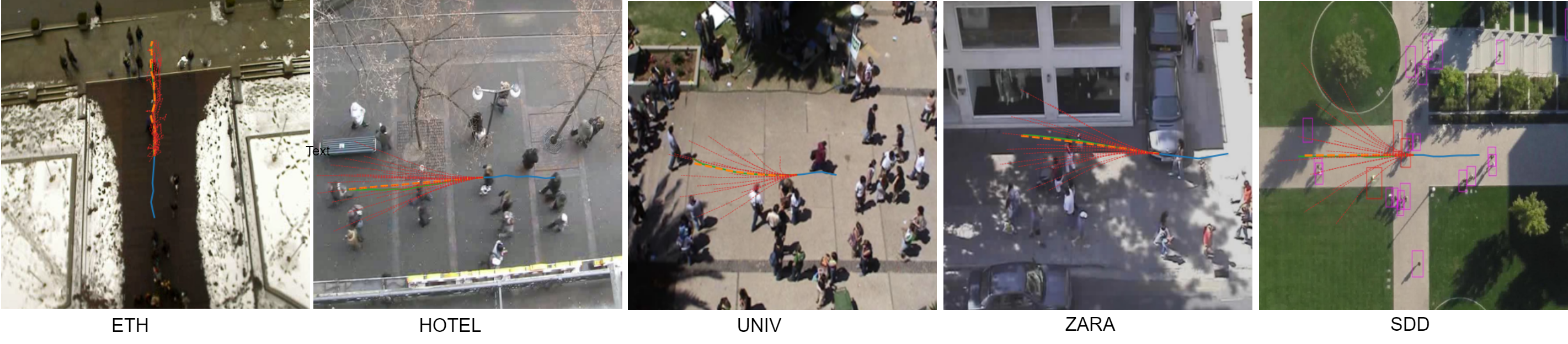}
 \caption{Illustration of the predicted pedestrian trajectories on the ETH-UCY and SDD datasets. Predicted pedestrian trajectories from our approach are depicted in \textcolor{orange}{orange}, observed trajectories in \textcolor{blue}{blue}, and ground truth trajectories in \textcolor{green}{green}. Our approach accurately predicts future trajectories.}
    \label{fig:visualisation}
\end{figure}

\subsection{Comparison with the State-of-the-Art}

\textbf{Quantitative Analysis.}
Quantitative results on ETH-UCY and SDD datasets
are shown in Tables \ref{eth_main} and \ref{tab:sddmain}. We conducted a comparison of our CCF framework against state-of-the-art methods. The experimental results demonstrate the improved performance of our approach in terms of Average Displacement Error (ADE) and Final Displacement Error (FDE) on both the ETH-UCY and SDD datasets. In the ETH-UCY dataset, our approach achieves the best ADE results for HOTEL, UNIV, ZARA1, ZARA2, and the second best for ETH. In terms of FDE, our approach achieves the best results for ETH, HOTEL, ZARA2, and the second best for UNIV and ZARA1. 
Specifically, CCF achieves an average ADE/FDE of 0.20/0.34 on the ETH-UCY dataset which are the best results among all compared methods. On the SDD dataset, our approach achieves the best results for both ADE and FDE.


\textbf{ Qualitative Analysis.} As depicted in Fig.~\ref{fig:visualisation}, our method accurately forecasts future trajectories, closely matching the ground truth. The ground truth future trajectory is shown in green, the predicted trajectory in orange, and the observed past trajectory in blue. For better visualization, we only show the trajectory for one pedestrian. The model trained using our approach successfully captures the spatio-temporal representations and predicts the accurate trajectory. Out of 20 sample trajectories (red colored), the best trajectory of the pedestrian is selected as the final predicted trajectory.

\begin{table}[!t]
\caption{(a) Reported results regarding the effectiveness of various components of CCF on the ETH-UCY dataset demonstrate that the inclusion of secondary task significantly improves trajectory prediction performance. Furthermore, the cross-correction mechanism enhances representation learning further, as evidenced by performance improvement. (b) Reported results on the ETH dataset regarding the selection of the Huber loss and the cross-entropy loss for the prediction and classification of future trajectories indicate that these losses yield accurate future prediction.}
\label{tab:my-table2}
\addtolength{\tabcolsep}{-1.5mm}
\resizebox{\textwidth}{!}{%
\begin{tabular}{cc} 
    \begin{tabular}{|ccc|c|c|}
    \hline
    \multicolumn{3}{|c|}{Variant}                                                        & \multirow{2}{*}{ADE} & \multirow{2}{*}{FDE} \\ \cline{1-3}
    \multicolumn{1}{|c|}{Transformer} & \multicolumn{1}{c|}{Secondary Task} & Cross Correction &                      &                      \\ \hline
    \multicolumn{1}{|c|}{\checkmark}  & \multicolumn{1}{c|}{-}       & -      &  0.45                    &    0.85                  \\ \hline
    \multicolumn{1}{|c|}{\checkmark}  & \multicolumn{1}{c|}{\checkmark}   & -       &            \underline{0.23}        &     \underline{0.37}                   \\ \hline
    \multicolumn{1}{|c|}{\checkmark} & \multicolumn{1}{c|}{\checkmark} & \checkmark & \multicolumn{1}{l|}{\textbf{0.20}} & \multicolumn{1}{l|}{\textbf{0.34}} \\ \hline
    \end{tabular} 
    &
    \begin{tabular}{|c|c|c|c|}
    \hline
    Task                            & Variant       & ADE & FDE \\ \hline
    \multirow{2}{*}{Primary Task}     & MSE Loss          &  \underline{0.40}   &  \underline{0.59}   \\ \cline{2-4} 
                                    & Huber Loss       &   \textbf{0.35}  &   \textbf{0.55}  \\ \hline
    \multirow{2}{*}{Secondary Task} & BCE (one-hot) w Sigmoid & 0.39    &  0.58  \\ \cline{2-4} 
                                    & CE (probability) w Softmax  &  \textbf{ 0.35}  & \textbf{0.55}     \\ \hline
    \end{tabular}
\end{tabular}%
}
\end{table}

\begin{table}[!t]
\caption{(a) Reported results regarding the choice of the DNet for generating transformed/diverse observed trajectories. Results indicate that the output of the DNet network does not exactly replicate the original past observed trajectory but instead generates a diverse version of it. (b) Comparison of DNet with other approaches to incorporate diversity into the learning process. It is evident from the results that the DNet provides better diversity compared to the other methods.}
\label{tab:diverseNet}
\addtolength{\tabcolsep}{-1.2mm}
\resizebox{\textwidth}{!}{%
\begin{tabular}{cc}
    \begin{tabular}{|c|c|c|c|c|}
    \hline
    Method                      & \multicolumn{1}{c|}{Variant} & \multicolumn{1}{c|}{MSA} & MAE & \multicolumn{1}{c|}{RMSE} \\ \hline
    \multirow{6}{*}{DNet} & \multicolumn{1}{c|}{ETH}     & \multicolumn{1}{c|}{2.59}    &   0.80  & \multicolumn{1}{c|}{1.61}     \\ \cline{2-5} 
                                & \multicolumn{1}{c|}{HOTEL}   & \multicolumn{1}{c|}{0.79}    &  0.47   & \multicolumn{1}{c|}{0.89}     \\ \cline{2-5} 
                                & UNIV                         & 0.92                          &  0.50  &     0.96                       \\ \cline{2-5} 
                                & ZARA1                        & 0.92                         & 0.50    &         0.96                  \\ \cline{2-5} 
                                & ZARA2                        & 0.74                         & 0.45    &     0.86                      \\  \hline
    \end{tabular}
    &
    \begin{tabular}{|c|c|c|c|c|c|c|}
    \hline
    Methods    & ETH & HOTEL & UNIV & ZARA1 & ZARA2 & AVG \\ \hline
    Noise    & 0.43/0.62    & 0.11/0.17      & 0.24/0.43     &      0.20/0.37 &     0.15/0.27  &  0.23/0.38  \\ \hline
    Drop    &  0.44/0.64   &    0.12/0.18   &    0.23/0.43  &    0.19/0.35   &  0.14/0.25     &  0.23/0.37   \\ \hline
    Mask    & 0.47/0.64    &      0.12/0.17 & 0.24/0.44      &   0.19/0.35    &    0.14/0.26   &  0.23/0.37   \\ \hline
    DNet &   0.38/0.55  &   0.11/0.17    &   0.23/0.42   &    0.18/0.34   &  0.14/0.25     & \textbf{ 0.20/0.34}   \\ \hline
   
    \end{tabular}
\end{tabular}%
}
\end{table}

\subsection{Ablation Studies.} \label{sec:abl}
\subsubsection{Effectiveness of Various Modules Used in Our Approach}
\label{sec:componets}
We conducted an ablation study to validate the effectiveness of different components of our approach. Table \ref{tab:my-table2} (a) shows that the inclusion of a secondary task improves performance from 0.45/0.85 to 0.23/0.37, validating that the secondary task aids the primary task. The results also show that including the proposed cross-correction improves trajectory prediction performance from 0.23/0.37 to 0.20/0.34, representing a relative percentage gain of 14\%/9\% in average (AVG) ADE/FDE on the ETH-UCY dataset. 

\subsubsection{Results Using Different Losses for Primary and Secondary Tasks} 
\label{abl:loss}

We conducted a study to evaluate different loss functions for training our CCF model on the ETH dataset. The results, reported in Table \ref{tab:my-table2} (b), include MSE loss and Huber Loss for the primary task of trajectory prediction, and cross-entropy loss (with softmax) and binary cross-entropy loss (with sigmoid) for the secondary task of predicting class probabilities. For the primary task, Huber loss achieved better trajectory prediction performance with ADE/FDE values of 0.35/0.55, which is an improvement over the MSE results of 0.40/0.59. For the secondary task, cross-entropy loss achieved better performance with ADE/FDE values of 0.35/0.55, surpassing the 0.39/0.58 achieved with binary cross-entropy loss.


\subsubsection{Comparison Among Different Methods to Introduce Diversity}
\label{sec:diversity}
We assess the effectiveness of DNet in inducing diversity among past trajectories on ETH-UCY. From Table \ref{tab:diverseNet} (a), it is evident that the DNet generates a diverse trajectory, as shown by the average MSE/MAE/RMSE errors between the original past observed trajectory and its transformed version over complete training data. Table \ref{tab:diverseNet} (b) shows a comparison of DNet with other methods, such as dropping, adding noise, and masking past observed trajectories to incorporate diversity into the learning process. The introduction of DNet in the CCF model significantly improves the trajectory prediction performance with average (AVG) ADE/FDE values of 0.20/0.34, which are comparatively better than other methods on the ETH-UCY dataset.


\section{Conclusion}
In this work, we propose a novel approach called the cross-correcting framework for trajectory prediction, which aims to better learn the spatio-temporal representations of pedestrian trajectories. It comprises a cross-correcting training framework in which two trajectory prediction models, known as subnets, are trained with cross-correction. This allows both subnets to correct each other. To induce diverse learning between the subnets, we use the DNet network, which induces diversity by transforming the observed trajectory. Our approach utilizes a transformer-based architecture for trajectory prediction, consisting of an encoder that captures motion patterns and a decoder that captures social interactions among pedestrians for accurate future trajectory prediction. Experimental results demonstrate that incorporating cross-correction, DNet, and secondary task enhances trajectory prediction performance. We showcase the effectiveness of our approach on two widely used pedestrian trajectory prediction benchmarks and present various ablation experiments to validate our approach.

\bibliographystyle{unsrt}
\bibliography{main}


\end{document}